%% file: samplepaper.tex
\documentclass[runningheads]{llncs}
\usepackage{graphicx}
\usepackage{booktabs} % For formal tables
\usepackage{multirow}
\usepackage{bm}
\usepackage{subfigure}
\usepackage{graphicx}
\usepackage{dblfloatfix}
\usepackage{amsmath}
\usepackage{amsfonts}
\usepackage{adjustbox}

\begin{document}
\title{Comparative Analysis of CNN-based Spatiotemporal Reasoning in Videos}

\author{Okan K\"op\"ukl\"u \and Fabian Herzog \and Gerhard Rigoll}
\institute{Institute for Human-Machine Communication\\Technical University Munich, Germany}

\maketitle           
\input{Tex/Abstract}
%
%
\input{Tex/Introduction}
\input{Tex/Related_Work}

\input{Tex/Methodology}

\input{Tex/Experiments}
\input{Tex/Conclusion}
\input{Tex/Acknowledgements}

\bibliographystyle{splncs04}
\bibliography{egbib}

\end{document}

%% file: Tex/Abstract.tex
\begin{abstract}
Understanding actions and gestures in video streams requires temporal reasoning of the spatial content from different time instants, i.e., spatiotemporal (ST) modeling. In this survey paper, we have made a comparative analysis of different ST~modeling techniques for action and gecture recognition tasks. Since Convolutional Neural Networks (CNNs) are proved to be an effective tool as a feature extractor for static images, we apply ST~modeling techniques on the features of static images from different time instants extracted by CNNs. All techniques are trained end-to-end together with a CNN feature extraction part and evaluated on two publicly available benchmarks: The Jester and the Something-Something datasets. The Jester dataset contains various dynamic and static hand gestures, whereas the Something-Something dataset contains actions of human-object interactions. The common characteristic of these two benchmarks is that the designed architectures need to capture the full temporal content of videos in order to correctly classify actions/gestures. Contrary to expectations, experimental results show that Recurrent Neural Network (RNN) based ST~modeling techniques yield inferior results compared to other techniques such as fully convolutional architectures. Codes and pretrained models of this work are publicly available\footnote{\url{https://github.com/fubel/stmodeling}}.  

\keywords{Spatiotemporal modeling, CNNs, RNNs, activity understanding, action/gesture recognition}
\end{abstract}

% Sparse temporal sampling strategy is actively used in many action and gesture recognition tasks requiring long term spatio-temporal modeling. Convolutional Neural Networks proved to be an effective tool as a feature extractor for static images. Yet, establishing the temporal relations of these extracted features plays a critical role on the overall performance of the constructed architecture. In this paper, we analyze and compare plenty of spatio-temporal modeling techniques. The performance of the applied techniques is tested on a publicly available Jester dataset. Experimental results shows that fully convolutional network with square kernels performs the best, multilayer perceptron (naive concatenation) performs the second best top1 accuracy. They prevail over all the other models, including numerous Recurrent Neural Network models.

%% file: Tex/Introduction.tex
\section{Introduction}

Deep learning has been successfully applied in the area of image processing, providing state of the art solutions for many of its problems such as super-resolution \cite{ledig2017photo}, image denoising \cite{Liu_2018_CVPR_Workshops}, and classification \cite{deng2009imagenet}. Due to the outstanding performance of two-dimensional (2D) Convolutional Neural Networks (CNNs) on processing static images,  many attempts have been made to generalize 2D CNN architectures to capture the spatiotemporal (ST) structure of videos \cite{Simonyan2014-vh}, \cite{Wang2016-gj}. Until recently, 2D CNNs were the only options for video analysis tasks since lack of large scale video datasets made it impossible to train 3D CNNs properly.

\begin{figure}[t!]
	\centering
	\includegraphics[width=0.7\columnwidth]{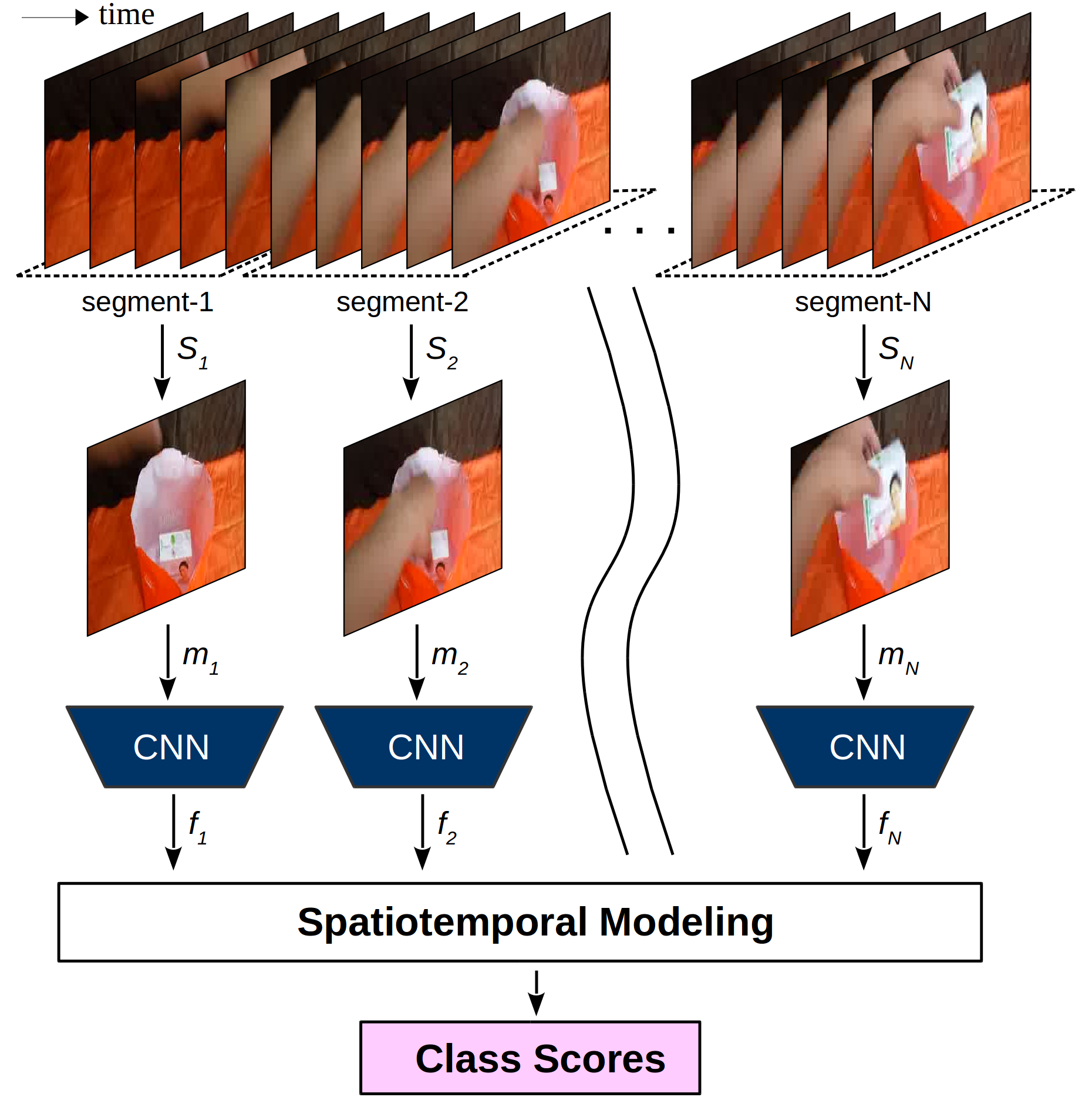}
	\caption{Spatio-Temporal Modeling Architecture: One input video containing an action/gesture is divided into \textit{N} segments. Afterwards, equidistant frames (m${_1}$,m${_2}$, .. m${_N}$) are selected from the segments and fed to a 2D CNN for feature extraction. Extracted features are fed to a ST~modeling block, which produces the final class score of the input video. In this example, action of ``taking something from somewhere" is depicted (a sequence from the Something-Something dataset).}
	\label{fig:STM_arch}
	\vspace{-0.3cm}
\end{figure}

With the availability of large scale video datasets such as Kinetics \cite{carreira2017quo}, deeper and wider 3D CNN architectures can be successfully trained to achieve better performance compared to 2D CNNs \cite{hara2018can}. More importantly, 3D CNNs can capture the ST patterns in videos inherently without requiring additional mechanisms. However, their drawback is that the input size should always remain the same for 3D CNNs such as 16 or 32 frames, which makes them not suitable for capturing temporally varying actions. This is not a problem for activity recognition tasks for Kinetics \cite{carreira2017quo} or UCF-101 \cite{soomro2012ucf101} datasets, as videos can be successfully classified using even very small snippets of the complete video. However, there are tasks where the designed architectures need to observe the complete video at once in order to make successful decisions. For these tasks, 2D CNN based architectures are still useful as the complete videos can be sparsely sampled with the desired number of segments and features of the selected frames can be extracted. Still, these architectures need an extra mechanism to provide ST~modeling of the extracted features. 

This work aims to analyze and compare various techniques for ST~modeling of the features extracted by a 2D CNN from sparsely sampled frames of action/gesture videos. Fig. \ref{fig:STM_arch} depicts the used ST~modeling architecture. A complete action/gesture video is divided into a predefined number of segments. From each segment, a frame is selected (randomly in training and equidistant in testing) and fed into the 2D CNN to extract its features. In order to understand which type of action/gesture is performed, an ST~modeling technique is used. In this work, we have analyzed multi-layer perceptron (MLP) based techniques such as simple MLP, Temporal Relational Network (TRN) and Temporal Segment Network (TSN), Recurrent Neural Network (RNN) based techniques such as vanilla RNN, gated recurrent unit (GRU), long short-term memory (LSTM),  bidirectional LSTM (B-LSTM) and convolutional LSTM (ConvLSTM) techniques, and finally fully convolutional network (FCN) techniques.

Although analyzed techniques have been used at several works in the literature, there has not been any comparative analysis to highlight the advantages of each ST~modeling technique. With this survey paper, we try to fill this gap by comparing each technique in terms of efficiency (i.e. number of parameters and floating-point operations) and classification accuracy. 

The proposed ST~modeling techniques are evaluated on two publicly available benchmarks: (i) The Jester dataset that contains dynamic and static hand gesture videos, (ii) the Something-Something dataset that contains videos of various human-object interactions. The common aspect of both these videos is that the proposed recognition architectures need to analyze the full content of the video in order to make a successful recognition, which makes them perfect benchmarks for analyzing ST~modeling techniques.

% The rest of the paper is organized as follows. In Section 2, we present related work in 2D CNN based action/gesture recognition. Section 3 explains the details of the analyzed ST~modeling techniques. Section 4 presents the experiments and results. Finally, Section 5 concludes the paper.  

%% file: Tex/Related_Work.tex
\section{Related Work}

Deep learning architectures for ST~modeling have been extensively studied in recent years, particularly in the context of action and gesture recognition \cite{Karpathy_undated-pv}, \cite{Simonyan2014-vh}, \cite{Zhou2018-ah}, \cite{Wang2016-gj}. Karpathy \textit{et al.} \cite{Karpathy_undated-pv} suggest several CNN architectures that fuse information across the temporal domain and applied the resulting models to the Spots-1M classification and UCF Action Recognition data sets. To speed up the training, they proposed a CNN-based multi-resolution architecture that could slightly improve the final results. Two stream CNNs \cite{Simonyan2014-vh}, \cite{feichtenhofer2016convolutional} fuse a spatial network processing the video frames with a temporal network using optical flow to obtain a common class score. These methods rely on separately processing the spatial and temporal components of the video, which can be a disadvantage. 3D convolutional neural networks, on the other hand, can be used to inherently learn the spatiotemporal structure of videos \cite{hara2018can}, \cite{kopuklu2019resource}. Tran \textit{et al.} \cite{Tran2015-lf} apply a 3D CNN architecture to obtain spatiotemporal feature volumes of input videos. To reduce training complexity, Sun \textit{et al.} \cite{Sun2015-va} propose a factorization of 3D spatiotemporal kernels into sequential 2D spatial kernels and separately handle sequence alignment. Although a sparse sampling strategy can be applied to the input value to span a larger time duration \cite{kopuklu2018analysis}, all 3D architectures have the disadvantage that the input size needs to be fixed, which limits their capability of handling data sets with varying video lengths. 

Recurrent neural networks are a natural choice for processing dynamic length video sequences, and several modern architectures have been proposed for action recognition in videos. LSTM \cite{hochreiter1997long} has been used in various video understanding tasks. Donahue \textit{et al.} \cite{Donahue2017-cf} employ an LSTM after CNN-based feature extraction on the individual frames to learn spatiotemporal components and apply the architecture on the UCF Action Recognition data set. Similarly, Baccouche \textit{et al.} \cite{Baccouche2011-pl} use 3D convolutional neural networks together with an LSTM network. Liu \textit{et al.} \cite{Liu2016-ls} suggest to modify the Vanilla LSTM architecture to learn spatiotemporal domains. GRU \cite{cho2014properties} is a popular variant of LSTM architecture which is actively used in video recognition tasks such as \cite{Dwibedi2018-hj}. There have been many other variants of LSTM architecture, which are summarized in \cite{greff2016lstm}. Another recurrent method is the Differentiable RNN \cite{veeriah2015differential} generated by salient motion patterns in consecutive video frames. 

Although LSTM structure is proven to be stable and powerful in modeling long range temporal relations in various studies \cite{sutskever2014sequence}, \cite{graves2013generating}, it handles spatiotemporal data using only full connections where no spatial information is encoded. ConvLSTM \cite{xingjian2015convolutional} addresses this problem by using convolutional structures in both the input-to-state and state-to-state transitions. ConvLSTM is first introduced for precipitation nowcasting task \cite{xingjian2015convolutional}, and later used for many other applications such as video saliency prediction \cite{song2018pyramid}, medical segmentation \cite{zhang2019spatio}.

Fully Convolutional Networks (FCN) is first proposed for image segmentation task \cite{long2015fully} and currently majority of segmentation architectures are based on FCNs. Later FCN architectures have been used at many other tasks such as object detection \cite{ren2015faster}, \cite{redmon2016you}. The idea of using convolution layers can also be applied to the task of ST modeling.

Methods like Temporal Segment Networks \cite{Wang2016-gj} enable processing longer videos by segmenting the input video into a certain number of segments, selecting short-length snippets randomly from each segment and finally fusing individual prediction scores. These prediction scores are the result of a spatial convolutional network operating on the samples frames and a temporal convolutional network operating on optical flow components. However, it must be noted that averaging is applied for the fusion of extracted features in TSN, which causes to lose the temporal order of the features. Similarly, Temporal Relation Networks \cite{Zhou2018-ah} extract a number of ordered frames from the input video, which are then passed through a convolutional neural network for feature extraction. Different from TSN, TRN keeps the order of the extracted features and tries to discover possible temporal relations at multiple time scales.

%% file: Tex/Methodology.tex
\vspace{-0.25cm}
\section{Methodology}

In this section, we first describe the complete ST~modeling architecture, which is based on a 2D CNN feature extraction part and one ST~modeling block. Afterward, we investigate different ST~modeling techniques in detail that can be used within this architecture. Finally, we will give the training details used in the experiments.

\subsection{ST~modeling Architecture}

As illustrated in Fig. \ref{fig:STM_arch}, a video clip \textit{V} that contains a complete action/gesture is divided into \textit{N} segments. Each segment is represented as $S_n \in \mathbb{R}^{w \times h \times c \times m}$ of $m \geq 1$ sequential frames with $224 \times 224$ spatial resolution and $c=3$ channels. RGB modality is used in all of the trainings. Afterward, within segments, equidistant frames are selected and passed to a 2D CNN model for feature extraction. Extracted features are first pooled and transformed to a fixed size of 256 (except for TSN where features are transformed to \textit{number-of-classes}) via a one-layer Multi-layer Perceptron (MLP) except for ConvLSTM and 3D-FCN techniques. For these two techniques, no pooling is applied at the feature extraction and number of channels is transformed to 256 by using a $1\times1$ 2D convolution layer.

For feature extraction, two different CNN models are used: (i) SqueezeNet \cite{iandola2016squeezenet} with simple bypass and (ii) Inception with Batch Normalization (BNInception) \cite{ioffe2015batch}. The reason to choose these models is that the performance of the investigated ST~modeling techniques can be evaluated with a lightweight CNN feature extractor (SqueezeNet) and relatively more complex and heavyweight CNN feature extractor (BNInception). In this way, \textit{CNN-model-agnostic} performance of evaluated techniques can be observed.  

Extracted features are finally fed to an ST~modeling block, which produces the final class scores of the input video clip. Next, we are going to investigate different ST~modeling techniques in detail that are used in this block.

\subsection{Multi-layer Perceptron (MLP) based Techniques}

MLP-based ST~modeling techniques are simple but effective to incorporate temporal information. These techniques make use of MLPs once or multiple times. Extracted features are then fed to these MLP-based ST~modeling blocks keeping their order intact. The intuition is that MLPs can capture the temporal information of the sequence inherently without knowing that it is a sequence at all.   

\subsubsection{Simple MLP} 

\begin{figure}[t!]
\begin{minipage}[b]{0.475\linewidth}
\centering
	\includegraphics[height=3cm]{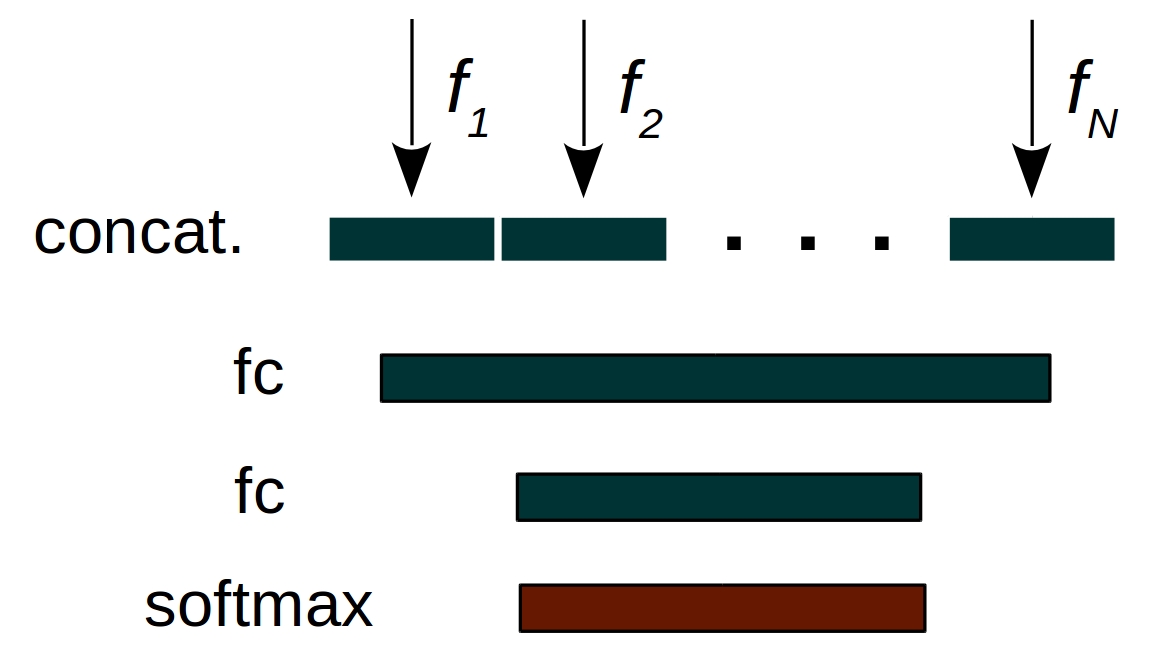}
\caption{Simple MLP technique. Extracted features are concatenated keeping their order same to form $N \time 256$ dimensional vector. This vector is fed to a 2-layer MLP to get final class scores.}
\label{fig:simple_MLP}
\end{minipage}
\hfill
\begin{minipage}[b]{0.475\linewidth}
\centering
    \includegraphics[height=3cm]{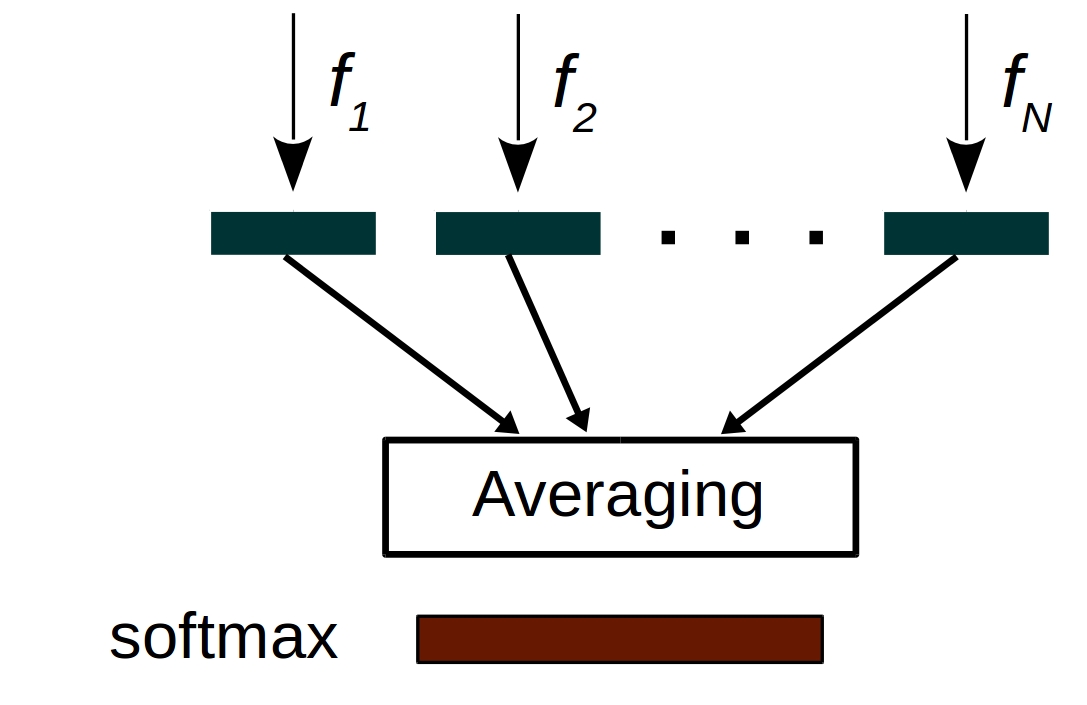}
\caption{Temporal Segment Network (TSN) architecture. Extracted frame features are transformed to \textit{Number-of-classes} dimension and averaged to get class conditional scores.}
\label{fig:TSN_arch}
\end{minipage}
\end{figure}

As illustrated in Fig. \ref{fig:simple_MLP}, extracted features are concatenated preserving their order. Then, the concatenated single $N \times 256$ dimensional vector is fed to a 2-layer MLP with 512 and \textit{Number-of-classes} neurons. Finally, the output is fed to a softmax layer to get class conditional scores. This is a simple but effective approach. Combined with other modalities such as optical flow, infrared and depth, competitive results can be achieved \cite{kopuklu2018motion}.

\subsubsection{Temporal Segment Network (TSN)} 

TSN aims to achieve long-range temporal structure modeling using sparse sampling strategy \cite{Wang2016-gj}. When the original paper was written, TSN achieved state-of-the-art performance on two activity recognition datasets, namely the UCF-101 \cite{soomro2012ucf101} dataset and the HMDB \cite{kuehne2011hmdb} dataset.

The original TSN architecture uses optical flow and RGB modalities, as well as different consensus methods such as evenly averaging, maximum, and weighted averaging. Among them, evenly averaging achieved the best results in the original experiments. Therefore, we have also experimented with evenly averaging for RGB modality only.

The corresponding TSN approach is depicted in Fig. \ref{fig:TSN_arch}. Unlike other ST~modeling techniques, the extracted frame features are transformed into a fixed size of \textit{number-of-classes} instead of 256. Afterward, all extracted features are averaged and fed to a softmax layer to get class conditional scores. 

Although TSN achieved state-of-the-art performance on UCF-101 and HMDB benchmarks at the time, it achieves inferior performance in the Jester and Something-Something benchmarks. The reason is that averaging causes loss of  temporal information. This does not create a huge problem for the UCF-101 and HMDB benchmarks as temporal order is not critical for these. Correct classification can even be achieved using only one frame of the complete video. However, the Jester and Something-Something datasets require the incorporation of the complete video in order to infer correct class scores.

\subsubsection{Temporal Relation Network (TRN)}

\begin{figure*}[t!]
	\centering
	\includegraphics[width=0.9\columnwidth]{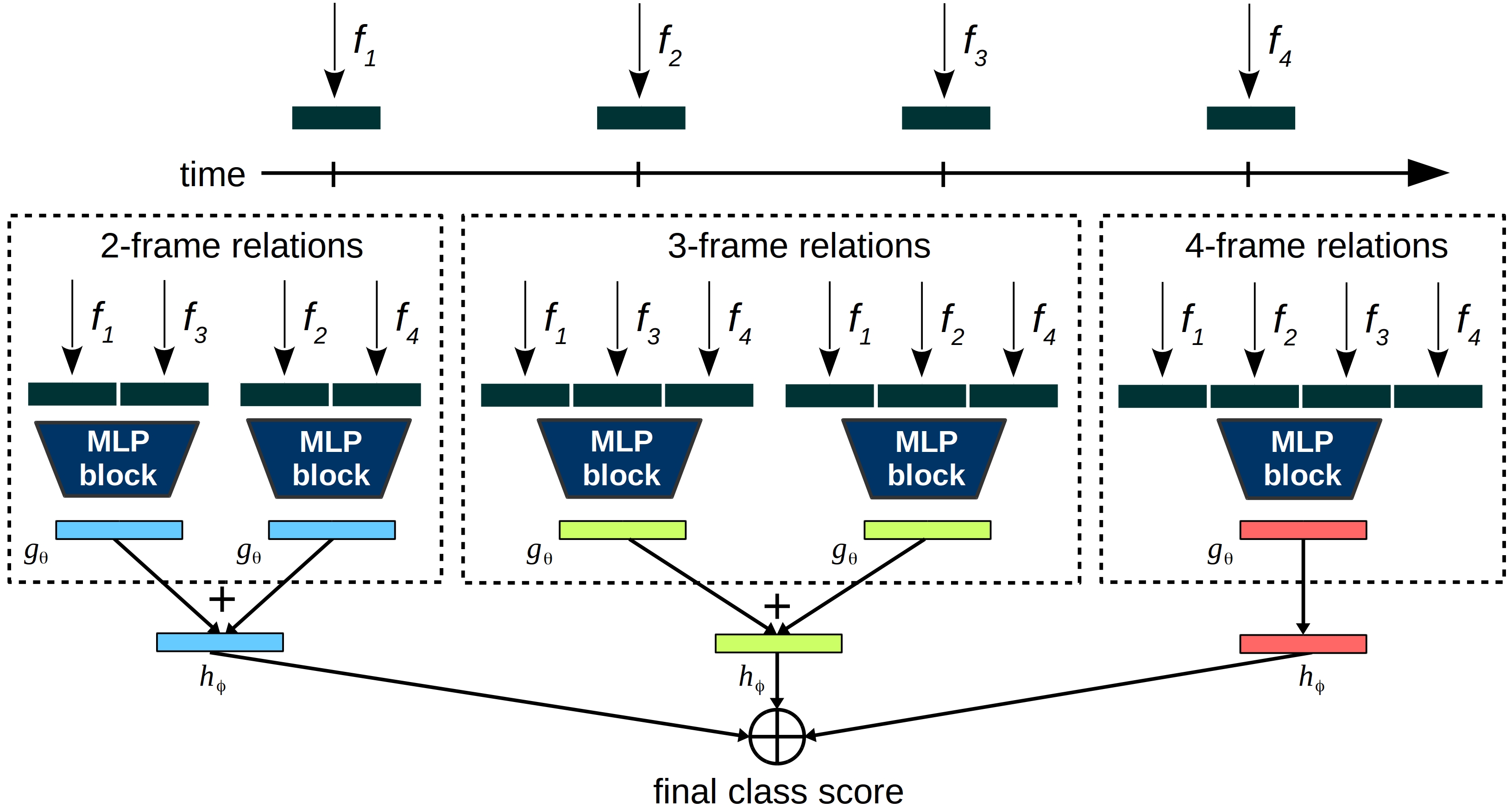}
	\caption{Illustration of Temporal Relation Networks. Features extracted from different segments of a video by a 2D CNN are fed into different frame relation modules. Only a subset of the 2-frame, 3-frame, and 4-frame relations are shown in this example (4 segments), as there are higher frame relations included according to the segment size.}
	\label{fig:TRN_arch}
\end{figure*}

TRNs \cite{Zhou2018-ah} aim to discover possible temporal relations between observations at multiple time scales. The main inspiration for this work comes from the relational reasoning module for visual question answering \cite{santoro2017simple}. The pairwise temporal relations (2-frame relations) on the observations of the video \textit{V} are defined as
\begin{align*}
    T_2(V) = h_{\phi} \left( \sum_{i<j}{g_{\theta}(f_i, f_j)} \right),
\end{align*}
\noindent where the input is the features of the $n$ selected frames of the video \textit{V} = \{$f_1$, $f_1$, ..., $f_n$\}, in which $f_i$ represents the feature of the $i^{th}$ frame segment extracted by a 2D CNN. Here, $h_\phi$ and $g_\theta$ represent the feature fusing functions, which are MLPs with parameters $\phi$ and $\theta$, respectively. For these functions, the exact same MLP block as depicted in Fig. \ref{fig:simple_MLP} is used. These two-frame temporal relations functions are further extended to higher frame relations, but the order of the segments should always be kept same in order to learn temporal relations inherently. Finally, all frame relations can be incorporated in order to get a single final output $MT_N(V) = T_2(V) + T_3(V) + ... + T_N(V)$, which is referred as multiscale TRN, where each $T_d$ captures temporal relationships between features of $d$ ordered frames. Fig.~\ref{fig:TRN_arch} depicts the overall TRN architecture.

% \begin{figure}[b!]
% 	\centering
% 	\includegraphics[width=0.7\columnwidth]{Figures/rnn_png1.png}
% 	\caption{M-layered architecture of Recurrent Neural Networks. }
% 	\label{fig:recurrent}
% \end{figure}

\subsection{Recurrent Neural Networks (RNN) based Techniques}
\begin{figure}[t!]
\begin{minipage}[b]{0.480\linewidth}
\centering
\includegraphics[height=3.5cm]{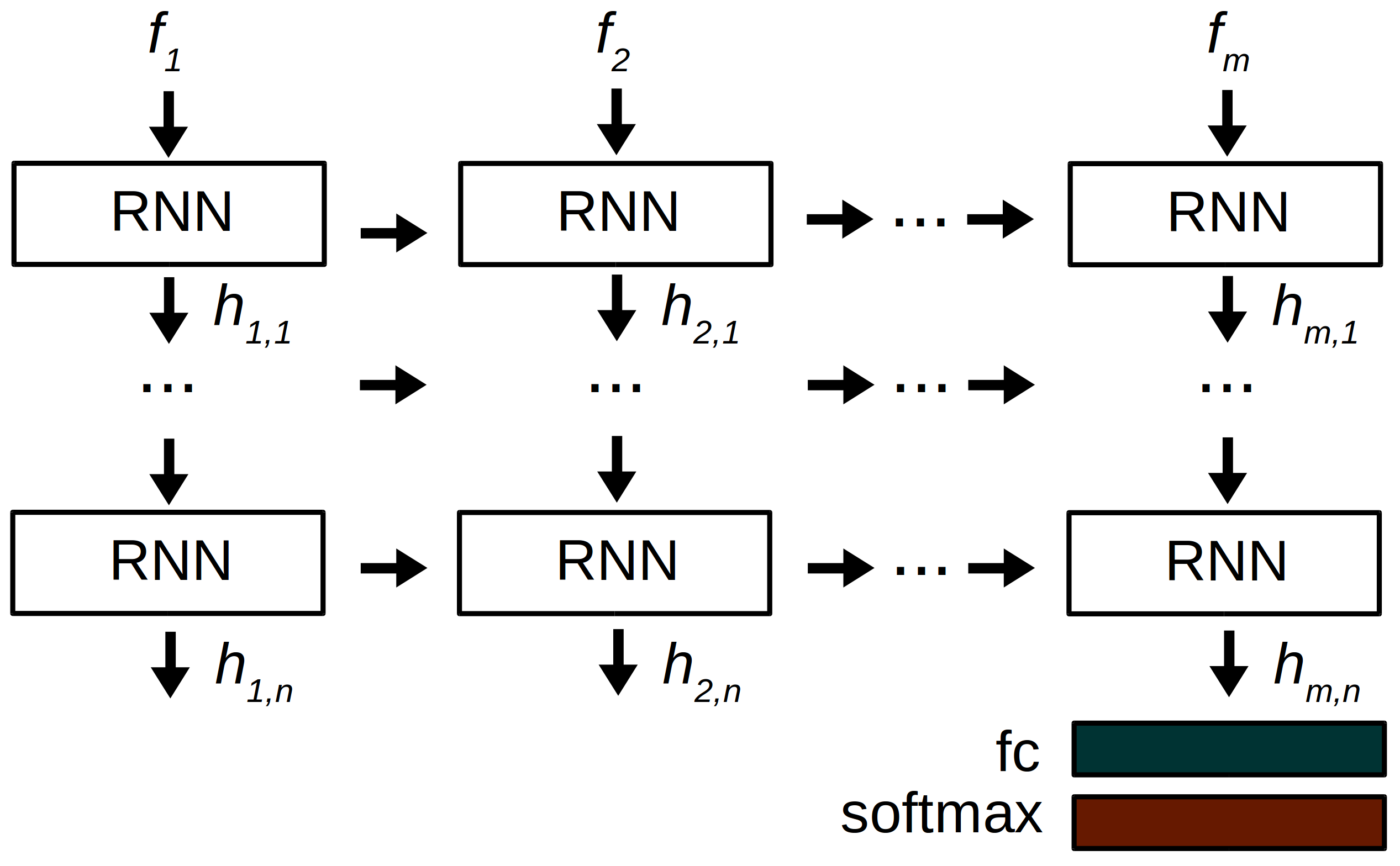}
	\caption{M-layered architecture of Recurrent Neural Networks.}
	\label{fig:recurrent}
\end{minipage}
\hfill
\begin{minipage}[b]{0.480\linewidth}
\centering
	\includegraphics[height=3.5cm]{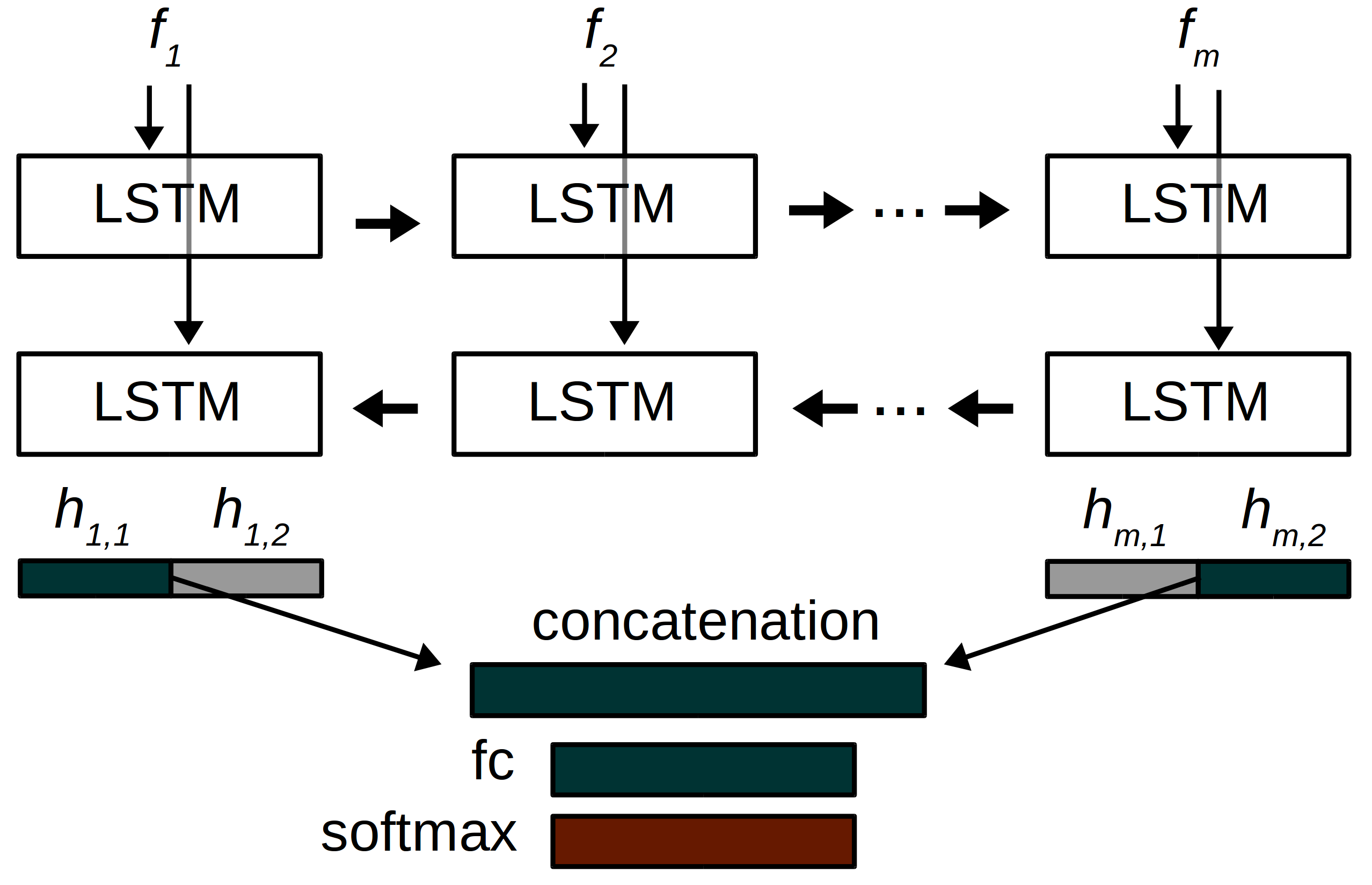}
	\caption{The data flow for Bidirectional LSTM architecture. .}
	\label{fig:bidirectional}
\end{minipage}
\end{figure}

Recurrent neural networks (RNNs) are a special type of artificial neural networks and consist of recurrently connected hidden layers which are capable of capturing temporal information. Furthermore, they allow the input and output sequences to vary in size. It is important to note that the hidden layer parameters do not depend on the time step but are shared across all RNN slices. The ability to keep information from previous time steps makes the hidden layer work like a memory. General M-layered RNN architecture is depicted in Fig. \ref{fig:recurrent}.

% \begin{figure*}[t!]
% 	\centering
% 	\subfigure[]{
% 	\includegraphics[width=0.35\linewidth]{Figures/rnns1.png}
% 	\label{fig:a}}
% 	\qquad
% 	\subfigure[]{
% 	\includegraphics[width=0.35\linewidth]{Figures/rnns2.png}
% 	\label{fig:b}}
% 	\qquad
% 	\subfigure[]{
% 	\includegraphics[width=0.35\linewidth]{Figures/rnns3.png}
% 	\label{fig:c}}
% 	\caption{Representation of the internal structure of Vanilla RNN (a), LSTM (b) and GRU (c).}%
% 	\label{fig:vanilla_rnn}
% \end{figure*}

%The shared parameters of an RNN can be learned by a method called \textit{backpropagation through time}. Theoretically, the hidden layer allows the network to learn any relations from the past. In practice, however, it turns out that classic RNNs suffer from two problems. By recursively forming derivatives, the gradient may vanish (vanishing gradient) or become too large (exploding gradient), which significantly limits the ability of classic RNNs.

In our experiments, we use two different vanilla RNNs, based on the hyperbolic tangent activation function, and the rectified linear unit (ReLU) activation function, respectively. Vanilla RNN with hyperbolic tangent activation function can be described by following equations
\begin{align*}
    \bm h_t &= \textbf{tanh}\left(\bm W_{hh}\bm h_{t-1} + \bm W_{xh}\bm x_t \right) \\
    \bm y_t &= \bm W_{hy}\bm h_t
\end{align*}
\indent Generally, we feed the output of the last node to a fully connected layer to obtain a vector size of the number of classes in the dataset. We also proceed in the same manner for all other RNN types except for the \textit{Bidirectional LSTM}. 

% \begin{figure}[b!]
% 	\centering
% 	\includegraphics[width=0.7\columnwidth]{Figures/blstm_full.png}
% 	\caption{The data flow for Bidirectional LSTM architecture. First and last laved outputs are concatenated and fed to a fully connected layer and softmax layer for classification.}
% 	\label{fig:bidirectional}
% \end{figure}

\subsubsection{Long Short-Term Memory (LSTM)}
LSTMs \cite{hochreiter1997long} are recurrent neural networks consisting of a cell, an input gate, a forget gate, and an output gate. The input gate $\bm i_t$ decides how much the current $\bm x_t$ contributes to the overall output. The cell $\bm c_t$ is responsible for remembering the previous state information, and also uses the results of the forget gate $\bm f_t$, which decides how much of the previous cell $\bm c_{t-1}$ flows into the current cell. 
As the name suggests, the forget gate can completely erase the previous state if necessary. Finally, the output gate determines the contribution of the current cell $\bm c_t$. All in all, the standard LSTM can be described by the following equations
\begin{align*}
    \bm i_t &= \bm \sigma \left( \bm W_{xi}\bm x_t + \bm W_{hi}\bm h_{t-1} + \bm W_{ci} \circ \bm c_{t-1} + \bm b_i \right) \\
    \bm f_t &= \bm \sigma \left( \bm W_{xf}\bm x_t + \bm W_{hf}\bm h_{t-1} + \bm W_{cf} \circ \bm c_{t-1} + \bm b_f \right) \\
    \bm c_t &= \bm f_t \circ \bm c_{t-1} + \bm i_t \circ \textbf{tanh}\left(\bm W_{xc}\bm x_t + \bm W_{hc}\bm h_{t-1} + \bm b_c \right) \\
    \bm o_t &= \bm \sigma \left( \bm W_{xo}\bm x_t + \bm W_{ho}\bm h_{t-1} + \bm W_{co} \circ \bm c_{t} + \bm b_o \right) \\
    \bm h_t &= \bm o_t \circ \textbf{tanh}\left(c_t \right)
\end{align*}
\noindent where `$\circ$' denotes the Hadamard product.

\subsubsection{Gated Recurrent Units (GRU)}
GRUs \cite{cho2014properties} are very similar to LSTMs and consist of two gates - an update gate $\bm z_t$ and a reset gate $\bm r_t$. However, unlike LSTMs, GRUs do not have their own memory control mechanism. Instead, the entire hidden layer information is directed to the next time step. The advantage of GRUs compared to LSTMs is their simplicity in structure, which significantly reduces the number of parameters to be learned. GRU can be described by the following qeuations
\begin{align*}
    \bm z_t &= \bm \sigma \left( \bm W_{xz}\bm x_t + \bm W_{hz}\bm h_{t-1} + \bm b_z \right) \\
    \bm r_t &= \bm \sigma \left( \bm W_{xr}\bm x_t + \bm W_{hr}\bm h_{t-1} + \bm b_r \right) \\
    \tilde{\bm h_t} &= \textbf{tanh}\left(\bm W_{xh}\bm x_t + \bm W_{hh}(\bm r_{t}\bm \circ \bm h_{t-1}) + \bm b_h \right) \\
    \bm h_t &= \bm z_t \circ \bm h_{t-1} + (1 - \bm z_t) \circ \tilde{\bm h_t}
\end{align*}
\noindent where `$\circ$' denotes the Hadamard product; $\tilde{\bm h_t}$ and $\bm h_t$ represent the intermediate memory and output, respectively.

\subsubsection{Bidirectional LSTM (BLSTM)}
BLSTMs \cite{schuster1997bidirectional} are a special form of LSTMs, but are trained in both directions. The fully connected layer is obtained by concatenating two halved outputs $\bm h_{1,1}$ and $\bm h_{m,2}$, namely the first output of the positive time direction and the last output of the negative time direction. The data flow for BLSTM architecture is depicted in Fig. \ref{fig:bidirectional}. We also investigate the effect of the hidden size by reducing it to half of the hidden size value we used for the other RNN-structures. This allows us to make meaningful comparisons with the latter. The reduction of the hidden layer size means that the vector size remains unchanged before the last fully connected layer. Consequently, the same number of output neurons is used for the classification.

\subsubsection{Convolutional LSTM (ConvLSTM)}
The main drawback of conventional LSTM (also GRU and vanilla RNN) in handling spatiotemporal data is that input-to-state and state-to-state transitions are made by full connections, where no spatial information is encoded. To overcome this drawback, convolutional LSTM proposes to use convolutional structures for the mentioned transitions. The main equations of ConvLSTM are given as
\begin{align*}
    \bm i_t &= \bm \sigma \left( \bm W_{xi} * \bm x_t + \bm W_{hi} * \bm h_{t-1} + \bm W_{ci} \circ \bm c_{t-1} + \bm b_i \right) \\
    \bm f_t &= \bm \sigma \left( \bm W_{xf} * \bm x_t + \bm W_{hf} * \bm h_{t-1} + \bm W_{cf} \circ \bm c_{t-1} + \bm b_f \right) \\
    \bm c_t &= \bm f_t \circ \bm c_{t-1} + \bm i_t \circ \textbf{tanh}\left(\bm W_{xc} * \bm x_t + \bm W_{hc} * \bm h_{t-1} + \bm b_c \right) \\
    \bm o_t &= \bm \sigma \left( \bm W_{xo} * \bm x_t + \bm W_{ho} * \bm h_{t-1} + \bm W_{co} \circ \bm c_{t} + \bm b_o \right) \\
    \bm h_t &= \bm o_t \circ \textbf{tanh}\left(c_t \right)
\end{align*}
\noindent where `$*$' and `$\circ$' denote the convolution operator and Hadamard product, respectively. In order to make use of ConvLSTM technique for ST~modeling, we have made some modifications. First, to keep spatial resolution, we have removed the final pooling layer of our feature extractor 2D CNN. Then, the output features are concatenated in time dimension forming a $D \times W \times H$ tensor. The last output of ConvLSTM is average pooled and fed to a final fully connected layer and a softmax layer to get class conditional scores.

\subsection{Fully Convolutional Network (FCN) based Techniques}
\label{fcnsk}

As the name implies, all of the layers of a fully convolutional network are convolutional layers. FCNs do not contain any linear (fully connected) layers at the end, which is the typical use for classification task. In order to utilize FCNs as ST~modeling technique, output features coming from the 2D CNNs can be concatenated over a dimension such that convolution operation can be performed over the concatenated tensor. If the features are pooled, concatenated features form a 2D tensor with over which 2D convolutional layers can operate. If pooling is not applied at feature extraction stage, concatenated features form a 2D tensor over which 3D convolutional layers can operate.

\begin{table}[t!]
	\centering
	\begin{tabular}{lcl}
		\specialrule{.15em}{.0em}{.2em}
		\textbf{Layer/Stride}    & \hspace{0.15cm} \textbf{Filter size} \hspace{0.15cm} & \textbf{Output size}\\ 
		\specialrule{.15em}{.2em}{.2em}
		Input                 &         & 1$\times$N$\times$256    \\
		Conv1/s(1,2)          & 3$\times$3     & 64$\times$N$\times$128   \\
		Conv2/s(1,2)          & 3$\times$3     & 64$\times$N$\times$64    \\
		Conv3/s(1,2)          & 3$\times$3     & 128$\times$N$\times$32   \\
		Conv4/s(1,2)          & 3$\times$3     & 128$\times$N$\times$16   \\
		Conv5/s(1,2)          & 3$\times$3     & 256$\times$N$\times$8    \\
	    \specialrule{.15em}{.2em}{.2em}
		Conv6/s(1,1)         & 1$\times$1   & \textit{NumCls}$\times$N$\times$8   \\
		AvgPool/s(1,1)        & N$\times$8   & \textit{NumCls} \\
		\specialrule{.15em}{.2em}{.3em}
	\end{tabular}
	\caption{Details of 2D fully convolutional ST~modeling architecture.}
	\label{tab:FCN2D_arch}
	\vspace{-0.3cm}
\end{table}

\begin{table}[t!]
	\centering
	\begin{tabular}{lcl}
		\specialrule{.15em}{.0em}{.2em}
		\textbf{Layer/Stride}    & \hspace{0.15cm} \textbf{Filter size} \hspace{0.15cm}  & \textbf{Output size}\\ 
		\specialrule{.15em}{.2em}{.2em}
		Input                 &         & 256$\times$D$\times$W$\times$H    \\
		Conv1/s(2,1,1)          & 3$\times$3$\times$3     & 64$\times$D/2$\times$W$\times$H     \\
		Conv2/s(2,1,1)          & 3$\times$3$\times$3     & 128$\times$D/4$\times$W$\times$H      \\
		Conv3/s(2,1,1)          & 3$\times$3$\times$3     & 256$\times$D/8$\times$W$\times$H     \\
	    \specialrule{.15em}{.2em}{.2em}
		Conv4/s(1,1,1)         & 1$\times$1$\times$1   & \textit{NumCls}$\times$D/8$\times$W$\times$H   \\
		AvgPool/s(1,1,1)        & D/8$\times$W$\times$H    & \textit{NumCls} \\
		\specialrule{.15em}{.2em}{.3em}
	\end{tabular}
	\caption{Details of 3D fully convolutional ST~modeling architecture.}
	\label{tab:FCN3D_arch}
	\vspace{-0.3cm}
\end{table}

\subsubsection{2D-FCN}

The inputs to 2D-FCN are the concatenated feature vectors of each segment resulting $N \times 256$ such that each row represents features from a segment. The input volume enters a series of 2D convolutions with stride $(1,2)$, which keeps the temporal dimension (i.e. the number of segments) intact throughout convolution operations. The kernel size is set to $3\times3$ with the same padding for all convolutions. After applying five convolutions, 2D convolution with $1\times1$ kernel is applied where the number of channels equals the number of classes. Finally, average pooling with $N\times8$ is applied to get class conditional scores. After each convolution, batch normalization and ReLU is applied. The details of the used 2D-FCN are given in \mbox{Table \ref{tab:FCN2D_arch}}.

\subsubsection{3D-FCN}

In order to make use of spatial information, similar to ConvLSTM, we have not used a pooling layer at the end of feature extractor 2D CNN, and output features are concatenated in depth dimension to create $D \times W \times H$ tensor. This tensor enters a series of 3D convolutions with stride $(2,1,1)$ in order to keep the spatial resolution same. The kernel size is set to $3\times3\times3$ with the same padding for all convolutions. After applying three convolutions, 3D convolution with $1\times1\times1$ kernel is applied to reduce the number of channels to number of classes, which is pooled later to get class scores. After each convolution, batch normalization and ReLU is applied. The details of the used 3D-FCN are given in \mbox{Table \ref{tab:FCN3D_arch}}. 

The proposed approach in fact very similar to rMCx models in \cite{tran2018closer} except for that 3D convolutions are applied at the very end. It is again similar to ECO architecture \cite{zolfaghari2018eco}, but 2D features in ECO architecture are extracted again at an early stage of the 2D CNN feature extractor. Although, 3D CNN architectures can be used for varying video lengths, we can use 3D convolution layers as ST~modeling technique since we are using a fixed segment size for all the input videos.

\subsection{Training Details}

Given the ST~modeling architecture in Fig. \ref{fig:STM_arch}, the CNN architecture used to extract frame features plays a critical role in the performance of the overall architecture. In order to get \textit{CNN-model-agnostic} performance of the applied ST~modeling techniques, the SqueezeNet and BNInception models are used. For both models, features are transformed to 256-dimensional vectors (\textit{Number-of-classes}-dimensional vectors for only TSN) via an MLP after global pooling layer except for ConvLSTM and FCN3D. For these two ST~modeling approach, spatial resolution ($13\times13$ and $7\times7$ for SqueezeNet and BNInception) is preserved by removing the final pooling layer, and $1\times1$ convolution layer is used to transform the number of channels to 256. For all experiments, CNN models pretrained on ImageNet dataset are used.  

\textbf{Learning:} Stochastic gradient descent (SGD) with standard categorical cross-entropy loss is applied. For momentum and weight decay, $9 \times 10^{-1}$ and $5 \times 10^{-4}$ are used, respectively. The learning rate is initialized with $1 \times 10^{-3}$ and reduced twice with a factor of $10^{-1}$ after validation loss converges.

\textbf{Regularization:} Several regularization techniques are applied in order to reduce over-fitting and achieve a better generalization. Weight decay of $\gamma = 5 \times 10^{-4}$ is applied to all parameters of the architecture. A dropout layer is applied after the global pooling layer of 2D CNN architectures with a ratio of $0.3$. Moreover, data augmentation of multiscale random cropping is applied for both datasets and random horizontal flip is applied for Something-Something dataset. Random horizontal flipping is not performed for the trainings of Jester dataset since this changes the annotations of some classes.  

\textbf{Implementation:} The complete ST~modeling architecture is implemented and trained (end-to-end) in PyTorch. % We make our code publicly available for reproducibility of the results\footnote{\url{https://github.com/fubel/stmodeling}}.

%% file: Tex/Experiments.tex
\section{Experiments}

\subsection{Datasets}

The Jester-V1 dataset is currently the largest hand gesture dataset that is publicly available \cite{jester}. It is an extensive collection of segmented video clips that contain humans performing pre-defined hand gestures in front of a laptop camera or webcam. The dataset consists of 148092 video clips under 27 classes, which is split into training, validation and test sets containing 118562, 14787 and 14743 videos, respectively. For the experiments, the validation set is used as the labels of the test set are not made available by dataset providers.

The Something-Something-V2 dataset is a collection of segmented video clips that show humans performing pre-defined basic actions with everyday objects \cite{Goyal_undated-kx}. It allows researchers to develop machine learning models capturing a fine-grained understanding of basic actions. The dataset consists of 220847 video clips under 174 classes, which is split into training, validation and test sets containing 168913, 24777 and 27157 videos, respectively. For the experiments, the validation set is used as the labels of the test set are not made available by dataset providers.

%The histograms for the duration of video clips are given in Fig. \ref{fig:jester_hist} and Fig. \ref{fig:sth_hist} for the datasets Jester and Something-Something, respectively. 
The duration of gesture clips in Jester dataset is concentrated between 30 - 40 frames. However, the Something-Something dataset has videos with relatively varying temporal dimension between 20 and 70 frames, which is the reason why 3D CNN architectures accepting fixed-size inputs are not suitable for this benchmark. In order to recognize video clips correctly, the used architectures should incorporate information coming from all parts of the videos.

\subsection{Resource Efficiency Analysis}

For real-time systems, the resource efficiency of the applied ST~modeling techniques is as essential as the achieved classification accuracy. Therefore, we have investigated the number of parameters and floating-point operations (FLOPs) of each technique, which can be found in Table \ref{tab:comparison}. For all calculations, we have used 8-segment case for Jester dataset. 

Out of all ST~modeling techniques, TSN comes for free since it requires no parameters and there is only averaging operation. However, temporal information is lost due to averaging, which results in inferior performance compared to simple-MLP or TRN techniques.

ConvLSTM and 3D-FCN does not employ pooling at the end of feature extractor, which requires highest number of FLOPs compared to others. The number of FLOPs for SqueezeNet is $13^2 / 7^2 = 3.48$ times higher than BNInception due to the output resolution of feature maps. In terms of number of parameters, ConvLSTM, 3D-FCN and TRN-multiscale requires the highest number with 4.73 M, 1.56 M and 2.34 M parameters, respectively. 

On the other hand, the resource efficiency of the feature extractors (i.e. 2D CNNs) are also important. The BNInception architecture contains 11.30 M parameters and requires 1894 MFLOPs to extract features of a \mbox{$224 \times 224$} frame. On the other hand, the SqueezeNet architecture contains only 1.24 M parameters and requires 338 MFLOPs to extract the features of a same-sized frame.

\begin{table*}[t!]
    \centering
    \begin{adjustbox}{width=1.0\textwidth}
    \begin{tabular}{lcccccccc}
        \specialrule{.15em}{.3em}{.3em}
        \multirow{3}{*}{\textbf{Model}} & \multirow{3}{*}{\textbf{MFLOPs}} & \multirow{3}{*}{\textbf{Params}} & \multicolumn{6}{c}{\textbf{Accuracy (\%)}}   \\ \cmidrule(lr){4-9}
         &   &   & \multicolumn{2}{c}{\textbf{Jester (8 seg.)}} & \multicolumn{2}{c}{\textbf{Something (8 seg.)}} & \multicolumn{2}{c}{\textbf{Something (16 seg.)}} \\ \cmidrule(lr){4-5} \cmidrule(lr){6-7} \cmidrule(lr){8-9}
         &   &   & \textbf{Squeez.} & \textbf{BNIncep.} & \textbf{Squeez.} & \textbf{BNIncep.} & \textbf{Squeez.} & \textbf{BNIncep.} \\
        \specialrule{.15em}{.3em}{.3em}
        % --------------------------------- % -------------------------------- % -----------------------%
        Simple-MLP      & 2.13   & 1.06M & 87.28  & 92.80  & 31.89  & 46.35  & 33.96  & 47.01 \\
        TSN             & 0.001  & 0.00M & 72.84  & 82.74    & 20.91  & 37.28    & 22.15  & 36.22 \\
        TRN-multiscale  & 11.95  & 2.34M & 88.39  & 93.20  & 33.73  & \textbf{46.91}  & 34.38  & \textbf{47.73} \\
        \specialrule{.15em}{.3em}{.3em}
        % --------------------------------- % -------------------------------- % -----------------------%
        RNN\_tanh       & 3.16   & 0.14M & 70.51  & 79.53  & 16.12  & 25.17  & 14.48  & 21.64 \\
        RNN\_ReLU       & 3.16   & 0.14M & 78.33  & 88.15  & 21.40  & 36.01  & 15.84  & 24.88 \\
        LSTM            & 8.42   & 0.53M & 84.28  & 90.80  & 25.24  & 39.04  & 28.25  & 42.83 \\
        GRU             & 6.32   & 0.40M & 83.10  & 90.86  & 25.40  & 40.69  & 30.24  & 43.31 \\
        B-LSTM          & 6.33   & 0.40M & 84.87  & 91.12  & 25.04  & 39.35  & 27.88  & 42.41 \\
        \specialrule{.15em}{.3em}{.3em}
        % --------------------------------- % -------------------------------- % -----------------------%
        ConvLSTM        & 1849.70$/$6379.54   & 4.73M   & 89.57    & 93.38    & 31.31    & 46.40    & 32.86    & 46.64 \\
        \specialrule{.15em}{.3em}{.3em}
        % --------------------------------- % -------------------------------- % -----------------------%
        2D-FCN       & 39.07  & 0.56M & 88.11  & 93.64  & 27.72  & 39.17  & 29.95    & 40.56 \\
        3D-FCN       & 152.07$/$524.49   & 1.56M   & \textbf{90.19}    & \textbf{94.07}    & \textbf{37.10}    & 46.66    & \textbf{37.59}    & 47.37 \\
        % \specialrule{.15em}{.3em}{.3em}
        % % % --------------------------------- % -------------------------------- % -----------------------%
        % DNDF            & ---   & ---   & ---    & ---    & ---    & ---    & ---    & --- \\
        % % \specialrule{.15em}{.3em}{.3em}
        % % % --------------------------------- % -------------------------------- % -----------------------%
        % % GMM-HMM(?)      & ---   & ---   & ---    & ---    & ---    & ---    & ---    & --- \\
        \specialrule{.15em}{.3em}{.3em}
    \end{tabular}
    \end{adjustbox}
    \caption{Comparison of different ST~modeling techniques over classification accuracy, number of parameters and computation complexity (i.e., number of Floating Point Operations - FLOPs). Methods are evaluated using 8 and 16 segments on validation sets of Jester-V1 and Something-Something-V2 datasets. The number of parameters and FLOPs are calculated for only ST~modeling blocks excluding CNN feature extractors for Jester dataset using 8 segments. FLOPs values of ConvLSTM and 3D-FCN are reported separately for BNInception (left) and SqueezeNet (right) since their spatial resolution is 7$\times$7 and 13$\times$13, respectively.}
    \vspace{-0.4cm}
	\label{tab:comparison}
\end{table*}

\subsection{Results Using Jester Dataset}

For the Jester dataset, the spatial content for all classes are the same: A hand in front of a camera performing a gesture. Therefore, a designed architecture should capture the form, position, and the motion of the hand in order to recognize the correct class.

Comparative results of different ST-modeling techniques for Jester dataset can be found in Table \ref{tab:comparison}. Inspired from \cite{kopuklu2018motion}, we have used eight segments for this benchmark as it achieves the best performance for MFF architecture. Compared to BNInception, architectures with SqueezeNet have 5-10\% inferior classification accuracy for the same ST~modeling technique. However, the technique-wise comparison remains similar within the same 2D CNN backbone.

Out of all ST~modeling techniques, TRN-multiscale, 2D-FCN, 3D-FCN and ConvLSTM stand out for classification accuracy. Considering the resource efficiency, the simple-MLP model can also be preferred over TRN-multiscale. Surprisingly, RNN-based methods except ConvLSTM, which first come to mind for modeling sequences, perform worse than these techniques. 

The superiority of ConvLSTM over other RNN based techniques and superiority of 3D-FCN over 2D-FCN validates the importance of the spatial content. 3D-FCN is the best performing technique in terms of accuracy for Jester dataset. However, preserving the spatial resolution brings the burden of increased computation and number of parameters. As expected, TSN yields the lowest classification accuracy as the averaging operation causes a loss of temporal information.

\subsection{Results Using Something-Something Dataset}

Compared to the Jester dataset, the Something-Something dataset contains much more classes with more complex spatial content. In order to identify the correct class label, the designed architectures need to extract the spatial content and temporally link this content successfully. Therefore, the frame feature extractors (i.e., 2D CNNs) are critical for the overall performance.

Comparative results of different ST-modeling techniques for the Something-Something dataset can be found in Table \ref{tab:comparison}. Beside 8-segment architectures, we have also made experiments for 16-segment architectures as the spatial complexity of the dataset is higher compared to Jester. Due to this complexity, architectures with SqueezeNet have 10\% to 15\% inferior classification accuracy compared to architectures with BNInception. However, similar to Jester dataset, the technique-wise comparison remains similar within the same 2D CNN backbone.

Compared to 8-segments, 16-segment architectures perform better. However, performance improvement is not as drastic as the effect of feature extractors. This shows that the main complexity of this task comes from the complexity of scenes, not the complexity of finer temporal details. In order to get better performance on Something-Something dataset, more complex architectures with deeper and wider structure can be preferred. 

Out of all ST~modeling techniques, 3D-FCN, ConvLSTM and TRN-multiscale again stand out for classification accuracy. Specifically, 3D-FCN with SqueezeNet performs best outperforming the second best model TRN-multiscale by 3.37\%. Similar performance difference cannot be observed when BNInception used as feature extractor. We conjecture that this is due to the higher feature resolution of the SqueezeNet architecture.

Moreover, ConvLSTM also outperforms all other RNN based techniques by 3-6\% showing the importance of spatial information again for this task. On the other hand, 2D-FCN cannot achieve the performance it reached in the Jester dataset and performs inferior to GRU and LSTM. Similar to the Jester dataset, Vanilla RNN and TSN yield lowest accuracies. 

%% file: Tex/Conclusion.tex
\section{Conclusion}

In this work, we have analyzed various techniques for CNN-based spatiotemporal modeling and compared them based on a consistent 2D CNN feature extraction of sparsely sampled frames. The individual methods were then evaluated on the Jester and Something-Something datasets. It has been shown that the CNN models used for feature extraction and the number of frames sampled affect the results. For the Jester dataset, the 3D-FCN technique achieves the best results using both SqueezeNet and BNInception. On the Something-Something dataset, again 3D-FCN and TRN-multiscale techniques outperform all other models while ConvLSTM performs similar results. It has also been shown that simple vanilla RNNs are unable to understand the complex spatiotemporal relationships of the data. All the more complex RNNs tested perform very similarly.

Interestingly, the TSN model, which showed state-of-the-art performance on the UCF-101 and HMDB benchmarks, performs rather poorly in our experiments, which shows the importance of maintaining the temporal information. Among all techniques, ConvLSTM and 3D-FCN requires the highest number of FLOPs, since they do not employ pooling at the feature extractor and preserve spatial resolution. For number of parameters, again ConvLSTM, 3D-FCN and TRN-multiscale techniques are the most expensive ones. While some models like TRN, LSTM, GRU, and B-LSTM can benefit from an increase in the number of segments, Vanilla RNNs and the TSN model can suffer from overfitting. One possibility for future research would be to develop resource efficient ST~modeling techniques that preserves the spatial resolution of the extracted features.

%% file: Tex/Acknowledgements.tex
\section*{Acknowledgements}
We gratefully acknowledge the support of NVIDIA Corporation with the donation of the Titan Xp GPU used for this research.